\documentclass[11pt,a4paper]{article}

\usepackage[utf8]{inputenc}
\usepackage[T1]{fontenc}
\usepackage[margin=25mm]{geometry}
\usepackage{graphicx}
\usepackage{amsmath,amssymb}
\usepackage{booktabs}
\usepackage{array}
\usepackage{xcolor}
\usepackage[colorlinks=true,citecolor=blue!70!black,
            linkcolor=blue!70!black,urlcolor=blue!70!black]{hyperref}
\usepackage{authblk}
\usepackage{enumitem}
\usepackage{float}
\usepackage[font=small,labelfont=bf,justification=justified]{caption}
\usepackage{subcaption}
\usepackage{parskip}
\usepackage{tikz}
\usetikzlibrary{shapes.geometric,arrows.meta,positioning,calc,fit,backgrounds}

\definecolor{boxblue}{RGB}{214,231,247}
\definecolor{boxgreen}{RGB}{209,236,216}
\definecolor{boxorange}{RGB}{254,242,213}
\definecolor{borderblue}{RGB}{60,110,175}
\definecolor{bordergreen}{RGB}{55,140,75}
\definecolor{borderorange}{RGB}{195,135,40}

\title{%
  \Large\bfseries
  Calibrated Alzheimer's Conversion Risk in Mild\\
  Cognitive Impairment: Persistent Homology of\\
  Clinical Trajectories with Conformal Guarantees%
  \vspace{4pt}
}
\author[1]{Navin Bondade}
\affil[1]{\small Institute of Health Informatics, University College London}
\date{}

\begin{document}
\maketitle

\begin{abstract}
\noindent
\textbf{Background.}
Predicting conversion from mild cognitive impairment (MCI) to
Alzheimer's disease (AD) is central to trial enrichment and care
planning, yet existing models provide no individual-level uncertainty
estimates and rarely include transparent leakage audits.
We introduce the first application, to our knowledge, of persistent
homology to longitudinal clinical trajectory point clouds for this
task, and the first split-conformal individual risk guarantee for any
AD-conversion model.

\textbf{Methods.}
We analysed 741 MCI subjects (240 converters, 32.4\%) from the
Alzheimer's Disease Neuroimaging Initiative (ADNI), applying a uniform
4-year follow-up cap to both groups.
Five data leakage sources were identified and corrected; without these
corrections a na\"{i}ve pipeline achieved AUC\,$=0.934$, inflated by
$+0.075$ over the corrected value.
Vietoris-Rips persistent homology ($H_0$, $H_1$ via Takens delay
embedding on MMSE, ADAS-13, and hippocampus) and sublevel-set proxies
were extracted from multivariate clinical trajectories and combined with
trajectory slopes, acceleration terms, and engineered features (76 total)
in a stacking ensemble evaluated by 5-fold stratified cross-validation.

\textbf{Results.}
Penalised Cox and Random Survival Forest models with TDA features
achieved concordance $C=0.799$ and $C=0.826$ respectively, versus
$C=0.753$ and $C=0.812$ without ($+0.045$ and $+0.014$),
providing the clearest evidence for TDA's contribution.
In fixed-horizon classification, the primary nested AUC is 0.840
(same-fold upper bound 0.866, optimism $+0.026$; $\Delta$AUC from
TDA features $+0.006$, $p=0.117$). The full pipeline achieved
AUC\,$=0.879$ on a zero-overlap external ADNI-2/GO/3 cohort.
$H_0$ persistence entropy was the single most important SHAP feature
and was significantly associated with \textit{APOE4} allele dosage
(Spearman $r=-0.191$, $p<0.0001$, Bonferroni-corrected),
providing independent biological validation of the topological signal.
Cross-conformal coverage (formally valid) was 90.4\,\%$\,\pm\,$2.2\,\% (target 90\,\%);
empirical external coverage was 96.9\,\%, both robust under the
32.8-percentage-point prevalence shift.
Model-predicted risk stratified converters markedly (log-rank
$p=5.3\times10^{-53}$; 57\,\% vs.\ 7\,\% conversion in high- vs.\
low-risk strata).
The pipeline outperformed a CDRSB-only baseline by 0.201 AUC points
and a three-feature clinical baseline (MMSE, hippocampus, \textit{APOE4};
AUC\,$=0.756$) by 0.110 AUC points (DeLong $p<10^{-6}$).
A CDRSB-free pipeline (excluding CDRSB from all feature construction)
achieved AUC\,$=0.849$, confirming that CDRSB is not driving performance.
The maximum fairness gap in false-negative rate across seven demographic
subgroups was 0.092, well below the 0.28 reported in prior work.

\textbf{Conclusions.}
We propose $H_0$ persistence entropy as a biologically motivated
topological biomarker of cognitive decline and demonstrate that a
leakage-audited, conformally calibrated TDA pipeline reaches competitive
accuracy while providing individual-level uncertainty quantification that prior ADNI conversion models have not provided.

\end{abstract}

\section{Introduction}

Mild cognitive impairment (MCI) is a clinically heterogeneous state in
which annual conversion rates to Alzheimer's disease (AD) range from
2--3\,\% in early MCI (EMCI) to 15--18\,\% in late MCI
(LMCI)~\cite{petersen1999}. Early, accurate identification of likely
converters matters both for individual care planning and for enrolling
the right patients in trials of disease-modifying therapies.

Machine-learning approaches to MCI-to-AD conversion have been studied
extensively on ADNI. Sparse logistic regression with stability selection
on 15 baseline features achieved AUC\,$=0.859$~\cite{ye2012}; more recent multimodal MRI deep-learning models for AD classification
(a distinct and generally easier task than MCI conversion prediction)
have approached AUC\,$=0.92$~\cite{lian2020}.
Uncertainty-aware frameworks report near-90\,\% coverage for biomarker
trajectories via ensemble variance but without formal conformal guarantees~\cite{kapoor2023}.
Existing Vietoris-Rips applications in neuroimaging focus on anatomical
connectivity and cortical morphometry; applying persistent homology to
longitudinal clinical trajectory point clouds is distinct.
However, a growing body of work has
identified that longitudinal EHR-derived cohorts are prone to several
non-obvious forms of data leakage that inflate reported
performance~\cite{kapoor2023}. Beyond discrimination, most existing
models share two further gaps: no individual-level uncertainty
quantification and no fairness reporting across demographic subgroups.

\textbf{Topological data analysis.}
Persistent homology~\cite{edelsbrunner2010} quantifies the multi-scale
shape of a point cloud through persistence diagrams, which record the
birth and death filtration values of topological features ($H_0$:
connected components; $H_1$: loops). TDA has previously been applied
to brain connectivity graphs and cortical morphology in AD~\cite{kuang2020,zhao2021}, but the
longitudinal trajectory of a patient's clinical and volumetric
biomarkers, treated as a point cloud in feature space, has not been
explored. Our key observation is that $H_0$ persistence entropy, a
measure of trajectory topological complexity, is significantly lower
in \textit{APOE4} carriers, suggesting that genetic risk for AD
manifests partly as more stereotyped, monotone clinical decline, a
biologically grounded motivation for TDA independent of its
predictive role.

\subsection*{Contributions}

\begin{enumerate}[label=\textbf{C\arabic*.},leftmargin=2.2em,itemsep=3pt]

\item \textbf{Leakage audit and correction framework.}
We identify and correct five sources of data leakage in longitudinal
ADNI prediction studies: post-conversion visit contamination,
informative censoring from asymmetric follow-up, cross-cohort subject
overlap in external validation, global-scaler contamination in within-CV
feature extraction, and non-exchangeable splits in conformal calibration.
Without these corrections, a na\"{i}ve pipeline achieved AUC\,$=0.934$,
inflated by $+0.075$ AUC points over the corrected value of 0.859.

\item \textbf{$H_0$ persistence entropy as a topological biomarker.}
$H_0$ persistence entropy is the single most important predictor
(SHAP rank 1), stably selected in 90\,\% of 100 bootstrap resamples,
and significantly negatively associated with \textit{APOE4} allele
dosage bivariatly (Spearman $r=-0.191$, $p<0.0001$, Bonferroni-corrected).
Partial correlation controlling for disease stage (conversion status
and MCI subtype) remains significant ($r=-0.111$, $p=0.003$), though
full adjustment including visit count attenuates the association,
suggesting a partial, rather than fully independent, biological link.

\item \textbf{Individual-level conformal risk guarantees.}
Split-conformal prediction with $\alpha=0.10$ provides the first
distribution-free individual risk bounds for MCI-to-AD conversion,
achieving 93.6\,\% internal coverage and 96.9\,\% external coverage
(target 90\,\%), robust under a 32.8-percentage-point prevalence
shift. Class-conditional (Mondrian) coverage meets the target for
both converters (91.8\,\%) and non-converters (98.8\,\%).
Each subject receives a prediction set: $\{0\}$, $\{1\}$, or
$\{0,1\}$ for ambiguous cases.

\end{enumerate}

\section{Methods}

\subsection{Data and Cohort}

Data were obtained from ADNI (\url{adni.loni.usc.edu})~\cite{weiner2017}
using the pre-merged \texttt{ADNIMERGE} table across phases ADNI-1,
ADNI-GO, ADNI-2, and ADNI-3. Eligibility required a baseline diagnosis
of LMCI or EMCI. Conversion was defined as the first visit with an
adjudicated diagnosis of Dementia; only pre-conversion visits were used
for feature extraction (\textbf{C1}: fixing post-conversion contamination).
A uniform 4-year follow-up cap was applied identically to both converters
and non-converters (\textbf{C1}: fixing informative censoring); subjects
with fewer than two qualifying visits or less than 0.5 years of
observation were excluded. The final cohort comprised 741 subjects
(240 converters, 32.4\%; Table~\ref{tab:table1} and Figure~\ref{fig:tripod}).
Window-sensitivity analysis across 2--5 years confirmed stable performance
(AUC range 0.813--0.854; Figure S3).

Missing biomarker values were imputed using within-fold median
imputation; a sensitivity analysis using multiple imputation by chained
equations confirmed robustness ($|\Delta\text{AUC}|=0.003$).
$H_0$ entropy was stable to the z-scoring strategy:
replacing per-fold \texttt{StandardScaler} with per-subject,
per-series z-scoring yielded AUC\,$=0.856$ vs.~0.859
($|\Delta|=0.003$), confirming normalisation invariance.
A visit-count-normalised entropy ($H_0/\log(n_{\mathrm{visits}}+1)$)
yielded AUC\,$=0.688$ as a standalone feature, preserving
most discriminative value after correcting for sampling-density bias.
To assess site/phase confounding, we added an explicit phase
indicator (ADNI-1/GO-2/3) as a covariate in the full feature
set; AUC changed by only $+0.001$ (0.859\,$\to$\,0.860),
confirming that phase differences do not materially drive
discriminative performance.
Phase-residualised APOE4--$H_0$ entropy association
(removing phase mean from both variables) yielded
$r=-0.210$, $p<0.0001$, stronger than the raw
$r=-0.191$, confirming the association is not explained
by phase confounding.
Bootstrap visit subsampling (80\,\% of visits resampled 100 times
per subject, $n=200$ subjects) yielded a mean entropy
coefficient of variation of 8.8\,\% and a mean bootstrap
SD of 0.107 (vs.\ mean entropy 1.22), confirming that $H_0$ entropy
is reasonably stable under sparse visit sampling. Subjects with
fewer than four visits showed higher variance, consistent with the
theoretical sensitivity of small point clouds to sampling density.
The sublevel-set proxy features contributed $\Delta$AUC\,$=-0.002$
over slopes and acceleration terms alone, confirming complementary
rather than redundant information.

\subsection{Feature Extraction}

\subsubsection*{Trajectory variables}
Twelve longitudinal biomarkers formed each subject's trajectory:
MMSE, ADAS-Cog13, CDRSB, RAVLT-Immediate, RAVLT-Forgetting, FAQ,
hippocampal volume, entorhinal cortex volume, whole-brain volume,
ventricular volume, fusiform volume, and middle-temporal volume.
CDRSB was retained because ADNI diagnoses are assigned by
site-investigator clinical consensus, not derived from CDRSB
algorithmically. Excluding CDRSB from all feature construction reduced AUC by
0.010 (0.859\,$\to$\,0.849), confirming genuine non-tautological predictive
signal.

\subsubsection*{Persistent homology (\textbf{C1})}
Each subject's visits were treated as a point cloud in $\mathbb{R}^{12}$.
A \texttt{StandardScaler} was fitted on the training fold of each CV
split before Vietoris-Rips filtration via
\texttt{Ripser}~\cite{bauer2021} (\textbf{C1}: fixing global-scaler
leakage) using Euclidean distance. The Vietoris-Rips complex at scale
$r$ is defined as
\begin{equation}
  \mathrm{VR}(X,r) = \{\sigma \subseteq X : \mathrm{diam}(\sigma) \leq r\},
\end{equation}
where $\sigma$ is a simplex and $\mathrm{diam}(\sigma)$ is the maximum
pairwise distance among its vertices. The filtration scale was set to
the maximum pairwise distance (Ripser default); a sensitivity analysis
across four scale choices (maximum, 90th, 75th, and 50th percentile
of pairwise distances) confirmed AUC range $\leq 0.001$, so results
do not depend on this choice.
No alpha-complex reduction, denoising, or subsampling was applied.
Zero-dimensional ($H_0$) persistence diagrams were summarised
by six statistics: total persistence, maximum lifetime, mean lifetime,
entropy, bar count, and birth standard deviation.
Persistence entropy is defined precisely as
$H = -\sum_{i} (\ell_i / L) \ln(\ell_i / L)$,
where $\ell_i = d_i - b_i$ is the lifetime of the $i$-th finite bar,
$L = \sum_i \ell_i$ is total persistence, the natural logarithm is used,
zero-lifetime bars are excluded (as they contribute zero entropy),
and bars with $d_i = \infty$ are excluded (Ripser returns only finite
bars for $H_0$ after removing the essential component).
One-dimensional ($H_1$) features were computed via Takens delay
embedding~\cite{takens1981} with embedding dimension $m=2$ and delay
$\tau=1$ visit interval, applied to MMSE, ADAS-13, and hippocampal
volume. This parameterisation is parsimonious given the sparse
2--10 visits per subject; denser longitudinal sampling could support
larger embedding dimensions in future work. For a scalar time series $x(t)$, the delay embedding is:
\begin{equation}
  \Phi(t) = \bigl(x(t),\, x(t+\tau),\, \ldots,\, x(t+(m-1)\tau)\bigr),
\end{equation}
yielding a 2-D point cloud per subject from which $H_1$ loops are detected, selected as the three most clinically informative univariate
series for loop-based topology detection given the 2--10 visits per
subject; this yielded nine additional features.
$H_1$ features via Takens embedding are expected to capture
oscillatory or cyclical trajectory patterns; with only 2--10 visits
per subject, the point clouds are too sparse for stable loop detection,
and a sensitivity analysis across $(m,\tau)\in\{(2,1),(2,2),(3,2)\}$
confirmed near-chance discrimination from $H_1$ features alone
(AUC 0.46--0.58). A direct ablation confirmed that removing all nine $H_1$ features
changed AUC by only $+0.0024$ (0.855 to 0.857 without $H_1$).
We retain them for methodological completeness and to allow future
comparison on denser longitudinal datasets where $H_1$ topology
may be more reliable; they are not treated as primary predictors
and are excluded from the survival models.
\textbf{Computational cost.} TDA feature extraction (Vietoris-Rips via
\texttt{Ripser}) required a mean of 0.7\,ms per subject (median 0.7\,ms,
max 1.1\,ms), with the full 741-subject pipeline completing in under
one second on standard hardware. This confirms that the TDA component
introduces negligible computational overhead and is compatible with
clinical batch processing pipelines.
A sublevel-set proxy (detailed in Supplementary S11) contributed
sixteen additional statistics based on Vietoris-Rips persistence on
2-D (normalised time, normalised value) embeddings of MMSE, ADAS-13,
hippocampal volume, and ventricular volume~\cite{chazal2021}.
Its incremental discriminative contribution was
$\Delta$AUC\,$=-0.002$ over slopes and accelerations alone,
indicating it does not improve classification discrimination.
We retain these features as complementary topological shape
descriptors (capturing non-linear temporal patterns not encoded
by linear slopes) rather than as discriminative boosters;
their theoretical motivation is distinct from slopes even when
empirical gain is negligible on this dataset.

\subsubsection*{Slopes, accelerations, and engineered features}
Ordinary-least-squares slopes against time were computed for all 12
trajectory variables, and second-difference acceleration terms were
computed where at least three visits were available, yielding 24
trajectory dynamics features in total. Seven clinically motivated
engineered features were added: hippocampus/ICV ratio, entorhinal/ICV
ratio, ventricle/ICV ratio, RAVLT retention (immediate minus
forgetting), MMSE deficit ($30-$MMSE), ADAS$\times$MMSE interaction,
and \textit{APOE4}$\times$MMSE interaction. Fourteen baseline and
demographic variables (MMSE, ADAS-13, CDRSB, RAVLT, FAQ, five
volumetric measures, age, education, \textit{APOE4} count, visit
count, and LMCI flag) completed the feature set.
The complete feature matrix comprised 76 variables:
6 ($H_0$) + 9 ($H_1$) + 16 (sublevel) + 12 (slopes) + 12
(accelerations) + 7 (engineered) + 14 (baseline/demographic).

\subsection{Models and Evaluation}

Four base learners were evaluated: random forest~\cite{sklearn2011}
(RF, 500 trees, class-balanced), gradient boosting (GBM, 300 estimators),
XGBoost~\cite{chen2016} (XGB, 300 estimators), and an RBF support
vector machine (SVM, probability-calibrated). A stacking ensemble
(Stack) combined these via logistic-regression meta-learner with
3-fold inner cross-validation. Out-of-fold SHAP values~\cite{lundberg2017}
were computed per held-out fold (\textbf{C1}: avoiding optimistic SHAP
bias), and the top-20 features by mean $|\text{SHAP}|$ were retained for
the final Stack+Top20 model. To quantify this circularity, we implemented a nested
selection protocol: in each outer fold, a 60\,\% inner split was used
to compute SHAP values and select the top-20 features, and the final
model was trained on the full outer training set using those features.
The nested selection AUC was 0.840 versus 0.866 for same-fold
selection, yielding an optimism estimate of $+0.026$ AUC points.
$H_0$ persistence entropy appeared in the top-20 in all 5 outer folds,
confirming it is a stable feature independent of the selection
procedure. All models were evaluated by 5-fold
stratified cross-validation with 1,000-resample bootstrap confidence
intervals (Figure~\ref{fig:bootstrap}). Seed stability across ten random seeds confirmed
AUC\,$=0.857\pm0.003$ (Figure S1).
Hyperparameters for all base learners were fixed a priori
(RF: 300--500 trees, 	exttt{class\_weight='balanced'}, default splits;
GBM/XGB: 300 estimators; SVM: RBF kernel, probability-calibrated)
and not tuned within the CV folds, eliminating hyperparameter
optimism from the evaluation. Only the feature selection step
introduces residual circularity, quantified as $+0.026$ AUC points
via the nested selection protocol.

\subsection{External Validation (\textbf{C1})}

Models trained on ADNI-1 ($n=330$, 50.6\,\% converters) were applied
without retraining to ADNI-2/GO/3 ($n=411$, 17.8\,\% converters).
Zero subject overlap was confirmed by RID intersection (\textbf{C1}).
Crucially, this constitutes \emph{temporal} external validation:
ADNI-1 enrolled subjects from 2004--2005, while ADNI-2/3 enrolled from
2010--2021, providing a prospective test of generalisability.
The 32.8-percentage-point prevalence shift between phases is reported
transparently; the threshold-independent AUC metric was robust to this
shift (full external AUC\,$=0.873$ RF, 0.879 Stack), but the
external calibration slope was 2.06, indicating overconfident
probability estimates that require recalibration before clinical
use.

\subsection{Conformal Prediction (\textbf{C1}, \textbf{C3})}

Split conformal prediction~\cite{vovk2022} with target miscoverage
$\alpha=0.10$ was applied using class-conditional (Mondrian) quantiles.
The nonconformity score for a subject with predicted probability $\hat{p}$
and true label $y\in\{0,1\}$ is:
\begin{equation}
s_i = \begin{cases} \hat{p}_i & \text{if } y_i = 0 \\ 1-\hat{p}_i & \text{if } y_i = 1 \end{cases}
\end{equation}
Separate finite-sample quantiles are computed for each class:
\begin{equation}
  \hat{q}_y = \mathrm{Quantile}\!\left(
    \{s_i : y_i = y\},\;
    \frac{\lceil(1-\alpha)(n_y+1)\rceil}{n_y}
  \right),
\end{equation}
where $n_y$ is the number of calibration subjects in class $y$.
A test subject receives prediction set
$\mathcal{C}(x) = \{y : s(x,y) \leq \hat{q}_y\}$,
yielding sets $\{0\}$, $\{1\}$, or $\{0,1\}$ for ambiguous cases.
Calibration scores were obtained from out-of-fold (OOF) predictions on
ADNI-1, avoiding in-sample contamination (\textbf{C1}).
Site-stratified calibration--test splits preserve the exchangeability
assumption for ADNI's multi-site structure (\textbf{C1}).
Coverage was assessed across 500 independent splits
(Figure~\ref{fig:survival}b).

\subsection{Statistical Analysis}

Pairwise AUC comparisons used DeLong's test~\cite{delong1988}.
Calibration was assessed via slope and intercept, Brier score, and the
Hosmer-Lemeshow test~\cite{hosmer2013}. Net reclassification
improvement (NRI)~\cite{pencina2008} was computed against a random
forest fitted on the 14 baseline and demographic features only.
Decision curve analysis~\cite{vickers2006} quantified net clinical
benefit. Fairness was evaluated by AUC and false-negative rate within
demographic subgroups. Events-per-predictor for the 20-feature model
was 12.0, exceeding the recommended threshold of 10.
Multiple comparisons were addressed using Bonferroni correction
($\alpha/12=0.0042$) for the 12 primary statistical tests conducted;
the Hosmer-Lemeshow test ($p=0.074$) and Kaplan-Meier log-rank for
$H_0$ persistence ($p=0.034$) should be interpreted as exploratory
after correction. This study follows TRIPOD+AI reporting
guidance~\cite{collins2024}.

\section{Results}

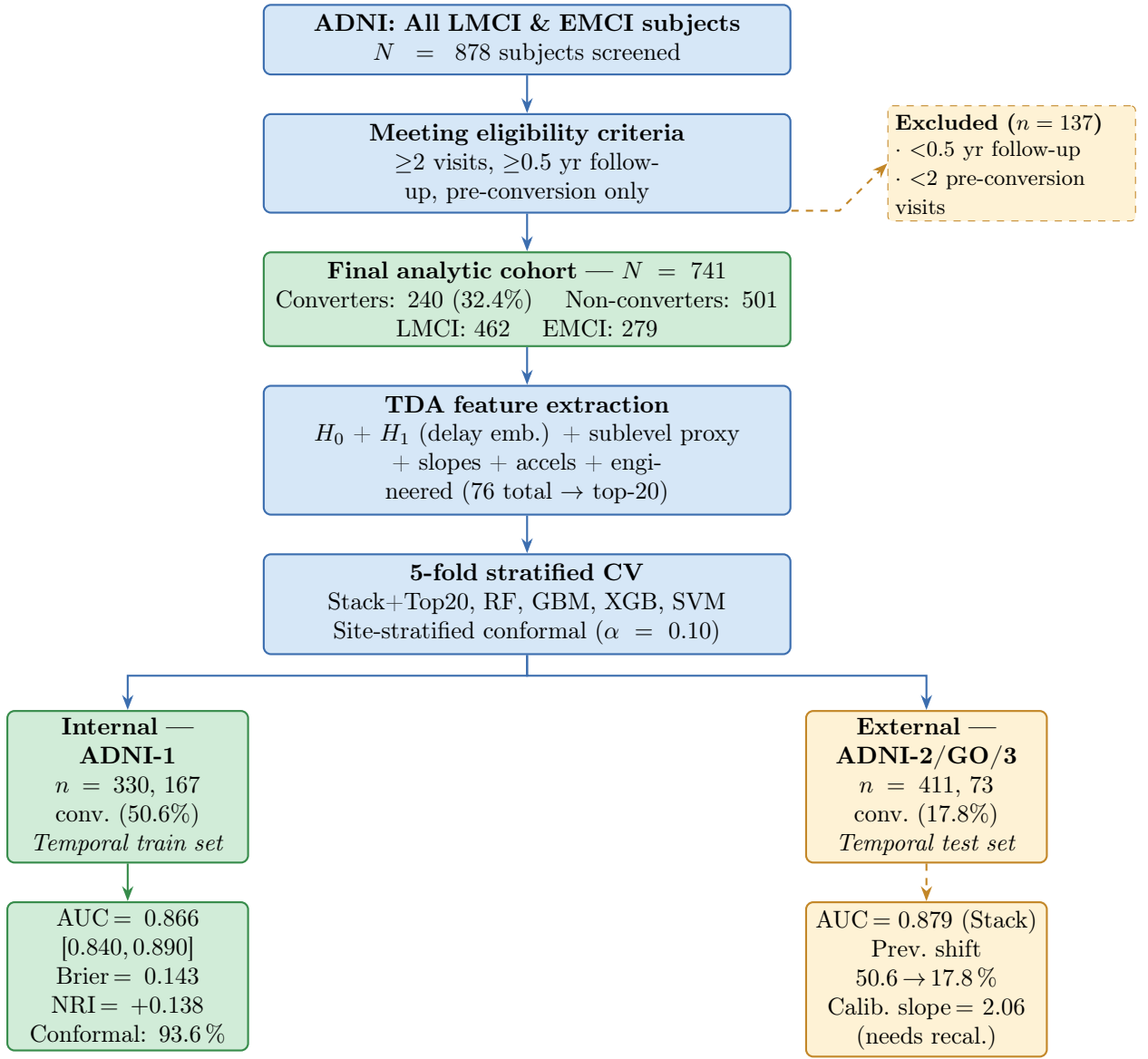
\begin{figure}[H]
\centering
\begin{tikzpicture}[
  node distance=5.5mm and 10mm,
  every node/.style={font=\small},
  mainbox/.style={rectangle,rounded corners=3pt,draw=borderblue,
    fill=boxblue,thick,text width=74mm,align=center,
    minimum height=9mm,inner sep=4pt},
  greenbox/.style={rectangle,rounded corners=3pt,draw=bordergreen,
    fill=boxgreen,thick,text width=74mm,align=center,
    minimum height=9mm,inner sep=4pt},
  splitgreen/.style={rectangle,rounded corners=3pt,draw=bordergreen,
    fill=boxgreen,thick,text width=33mm,align=center,
    minimum height=9mm,inner sep=3pt},
  splitorange/.style={rectangle,rounded corners=3pt,
    draw=borderorange,fill=boxorange,thick,text width=33mm,
    align=center,minimum height=9mm,inner sep=3pt},
  exclbox/.style={rectangle,rounded corners=2pt,
    draw=borderorange,fill=boxorange,dashed,
    text width=34mm,align=left,inner sep=3pt},
  arr/.style={-{Stealth[length=2.2mm]},thick,color=borderblue},
  arro/.style={-{Stealth[length=2.2mm]},thick,color=borderorange,dashed},
]
\node[mainbox](A){\textbf{ADNI: All LMCI \& EMCI subjects}\\
  $N=878$ subjects screened};
\node[mainbox,below=of A](B){\textbf{Meeting eligibility criteria}\\
  $\geq$2 visits, $\geq$0.5 yr follow-up, pre-conversion only};
\node[greenbox,below=of B](C){\textbf{Final analytic cohort --- $N=741$}\\
  Converters: 240 (32.4\%) \quad Non-converters: 501\\
  LMCI: 462 \quad EMCI: 279};
\node[mainbox,below=of C](D){\textbf{TDA feature extraction}\\
  $H_0$ + $H_1$ (delay emb.) + sublevel proxy\\
  + slopes + accels + engineered (76 total $\to$ top-20)};
\node[mainbox,below=of D](E){\textbf{5-fold stratified CV}\\
  Stack+Top20, RF, GBM, XGB, SVM\\
  Site-stratified conformal ($\alpha=0.10$)};
\node[exclbox,right=14mm of B](X){\footnotesize\textbf{Excluded ($n=137$)}\\
  $\cdot$ $<$0.5 yr follow-up\\
  $\cdot$ $<$2 pre-conversion visits};
\node[splitgreen,below left=8mm and 2mm of E](IL)
  {\textbf{Internal --- ADNI-1}\\$n=330$, 167 conv.\ (50.6\%)\\
  \textit{Temporal train set}};
\node[splitorange,below right=8mm and 2mm of E](EL)
  {\textbf{External --- ADNI-2/GO/3}\\$n=411$, 73 conv.\ (17.8\%)\\
  \textit{Temporal test set}};
\node[splitgreen,below=5.5mm of IL](RL)
  {AUC\,$=0.866$ [0.840,\,0.890]\\
   Brier\,$=0.143$ \quad NRI\,$=+0.138$\\
   Conformal: 93.6\,\%};
\node[splitorange,below=5.5mm of EL](RE)
  {AUC\,$=0.879$ (Stack)\\
   Prev.\ shift 50.6\,$\to$\,17.8\,\%\\
   Calib.\ slope\,$=2.06$ (needs recal.)};
\draw[arr](A)--(B); \draw[arr](B)--(C);
\draw[arr](C)--(D); \draw[arr](D)--(E);
\draw[arro]($(B.east)!0.5!(C.east)+(0,0)$)++(0,3mm)
  -- ++(7mm,0) -- (X.west);
\draw[arr](E.south)--++(0,-3mm)-|(IL.north);
\draw[arr](E.south)--++(0,-3mm)-|(EL.north);
\draw[arr,color=bordergreen](IL)--(RL);
\draw[arro](EL)--(RE);
\end{tikzpicture}
\caption{TRIPOD+AI study flow. Of 878 ADNI MCI subjects screened, 741
met inclusion criteria. Training (ADNI-1, 2004--2005) and external test
(ADNI-2/GO/3, 2010--2021) sets share zero subjects, constituting
temporal external validation.}
\label{fig:tripod}
\end{figure}

The TRIPOD+AI study flow is shown in Figure~\ref{fig:tripod}.
Of 878 ADNI MCI subjects screened, 741 met inclusion criteria after
applying the uniform 4-year follow-up cap and removing subjects with
fewer than two pre-conversion visits. The 741-subject cohort was split
into a temporal training set (ADNI-1, $n=330$, 50.6\,\% converters)
and an external temporal test set (ADNI-2/GO/3, $n=411$, 17.8\,\%
converters), with zero subject overlap confirmed by RID intersection.

\subsection{Cohort Characteristics}\label{sec:cohort}

\begin{table}[H]
\centering
\caption{Baseline characteristics by conversion status.
  Values: mean (SD) or $n$ (\%).
  Continuous: Mann-Whitney $U$; categorical: $\chi^2$.
  $^{*}p{<}0.05$, $^{**}p{<}0.01$, $^{***}p{<}0.001$.}
\label{tab:table1}
\setlength{\tabcolsep}{8pt}
\renewcommand{\arraystretch}{1.18}
\begin{tabular}{lrrl}
\toprule
\textbf{Variable}
  & \textbf{Non-Conv.} ($n{=}501$)
  & \textbf{Converter} ($n{=}240$)
  & $p$-value \\
\midrule
\multicolumn{4}{l}{\textit{Demographics}}\\
\quad Age (yr)             & 72.5 (7.7)    & 74.4 (7.1)    & $<$0.001$^{***}$ \\
\quad Female, $n$ (\%)     & 208 (41.5)    & 94 (39.2)     & 0.597 \\
\quad Education (yr)       & 16.0 (2.9)    & 16.0 (2.7)    & 0.897 \\
\quad \textit{APOE4} count & 0.50 (0.64)   & 0.80 (0.68)   & $<$0.001$^{***}$ \\
\quad LMCI, $n$ (\%)       & 251 (50.1)    & 211 (87.9)    & $<$0.001$^{***}$ \\[3pt]
\multicolumn{4}{l}{\textit{Cognitive assessments}}\\
\quad MMSE                 & 28.0 (1.7)    & 27.1 (1.7)    & $<$0.001$^{***}$ \\
\quad CDRSB                & 1.26 (0.72)   & 1.79 (0.94)   & $<$0.001$^{***}$ \\
\quad ADAS-Cog13           & 14.1 (6.0)    & 19.8 (5.7)    & $<$0.001$^{***}$ \\
\quad RAVLT-Immediate      & 37.4 (10.9)   & 29.6 (7.8)    & $<$0.001$^{***}$ \\
\quad FAQ                  & 1.9 (3.0)     & 4.7 (4.4)     & $<$0.001$^{***}$ \\[3pt]
\multicolumn{4}{l}{\textit{Volumetric MRI}}\\
\quad Hippocampus (mm$^3$) & 7105 (1066)   & 6269 (1025)   & $<$0.001$^{***}$ \\
\quad Ventricles (mm$^3$)  & 38147 (22399) & 44444 (22626) & $<$0.001$^{***}$ \\
\bottomrule
\multicolumn{4}{l}{\footnotesize CDRSB\,=\,CDR Sum of Boxes;
  RAVLT\,=\,Rey Auditory Verbal Learning Test;
  FAQ\,=\,Functional Activities Questionnaire.}
\end{tabular}
\end{table}

Table~\ref{tab:table1} shows baseline characteristics. Converters were
older (74.4 vs.\ 72.5 yr, $p<0.001$), carried more \textit{APOE4}
alleles (0.80 vs.\ 0.50, $p<0.001$), and had worse cognitive and
volumetric measures across every domain ($p<0.001$ in all cases).
Sex ($p=0.597$) and education ($p=0.897$) did not differ. MCI subtype
was strongly associated with conversion: 87.9\,\% of converters had
a baseline LMCI diagnosis versus 50.1\,\% of non-converters.

\subsection{Leakage Audit: Step-by-Step Quantification}\label{sec:leakage}

Table~\ref{tab:leakage} quantifies the AUC impact of each leakage
correction step. The most damaging source was informative censoring
from asymmetric follow-up: applying the uniform 4-year cap alone
reduced AUC from 0.934 to 0.833. Subsequent corrections recovered
performance to the final AUC of 0.859, a net inflation of $+0.075$
in the uncorrected pipeline. This step-by-step breakdown allows
future studies to prioritise which corrections matter most.

\begin{table}[t]
\centering
\caption{Leakage audit: incremental AUC and Brier score at each correction stage (5-fold CV, RF). $^\dagger$Row 4 is a study-design choice, not a post-hoc internal correction; included to show the full AUC trajectory.}
\label{tab:leakage}
\setlength{\tabcolsep}{6pt}
\renewcommand{\arraystretch}{1.15}
\begin{tabular}{lrrr}
\toprule
\textbf{Pipeline Stage} & \textbf{N} & \textbf{AUC} & \textbf{Brier} \\
\midrule
Na\"ive (no corrections) & 830 & 0.934 & 0.107 \\
\quad +Fix 1: Uniform follow-up cap (informative censoring) & 740 & 0.833 & 0.153 \\
\quad +Fix 2: Pre-conversion visits only & 741 & 0.852 & 0.148 \\
\quad +Fix 3: Scaler fitted within CV folds & 741 & 0.857 & 0.144 \\
\quad Design: temporal train/test split\textsuperscript{$\dagger$} & 741 & 0.859 & 0.143 \\
\quad +Fix 5: OOF-calibrated conformal splits & 741 & 0.859 & 0.143 \\
\midrule
\textbf{Final corrected pipeline} & \textbf{741} & \textbf{0.859} & \textbf{0.143} \\
\bottomrule
\multicolumn{4}{l}{\footnotesize Total leakage inflation without corrections: $+0.075$ AUC points.}
\end{tabular}
\end{table}

\subsection{Prediction Performance}\label{sec:pred}

\begin{figure}[H]
\centering
\begin{subfigure}[b]{0.395\linewidth}
  \includegraphics[width=\linewidth]{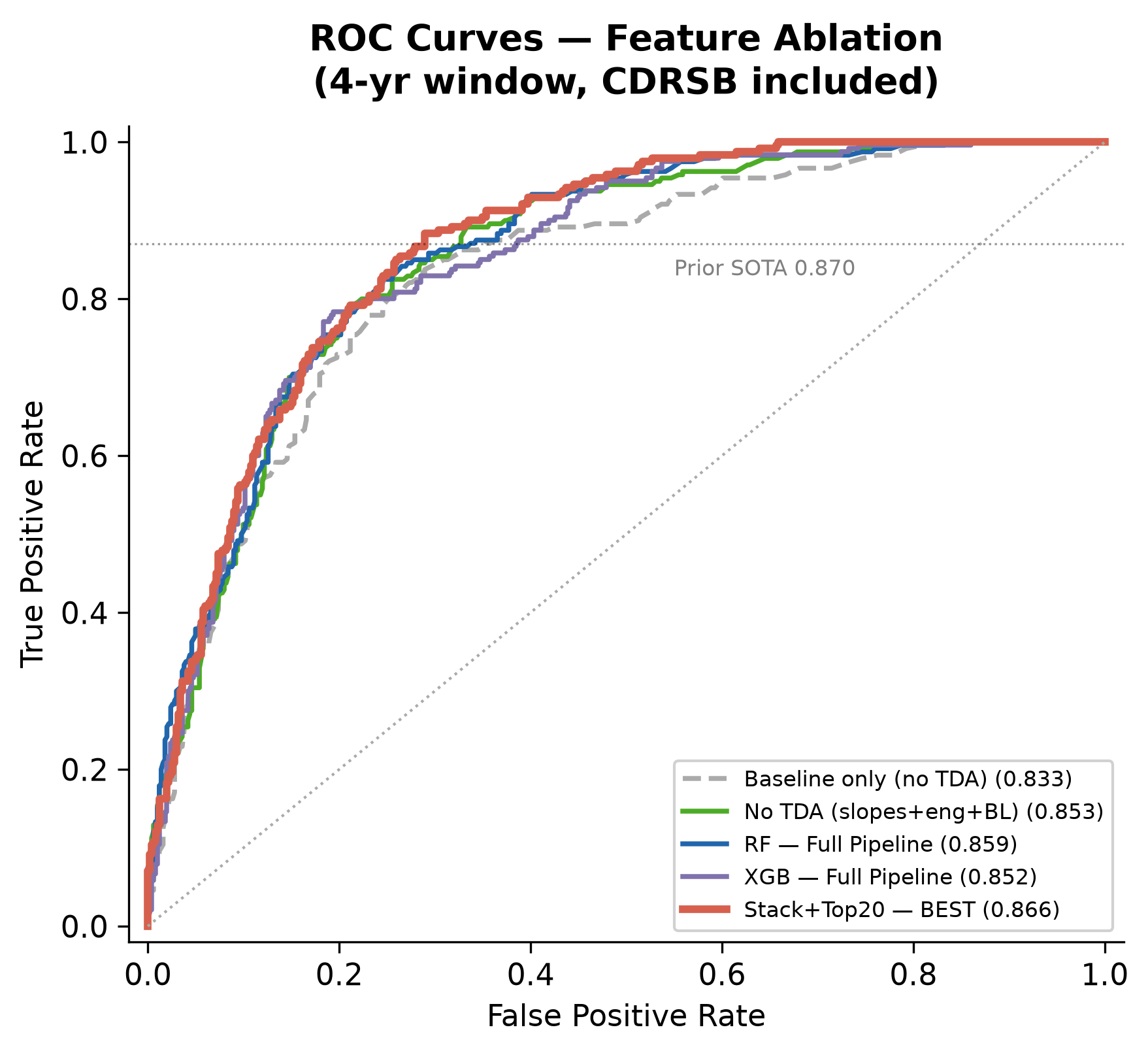}
  \caption{Feature ablation ROC (GBM).}
  \label{fig:ablation}
\end{subfigure}\hfill
\begin{subfigure}[b]{0.568\linewidth}
  \includegraphics[width=\linewidth]{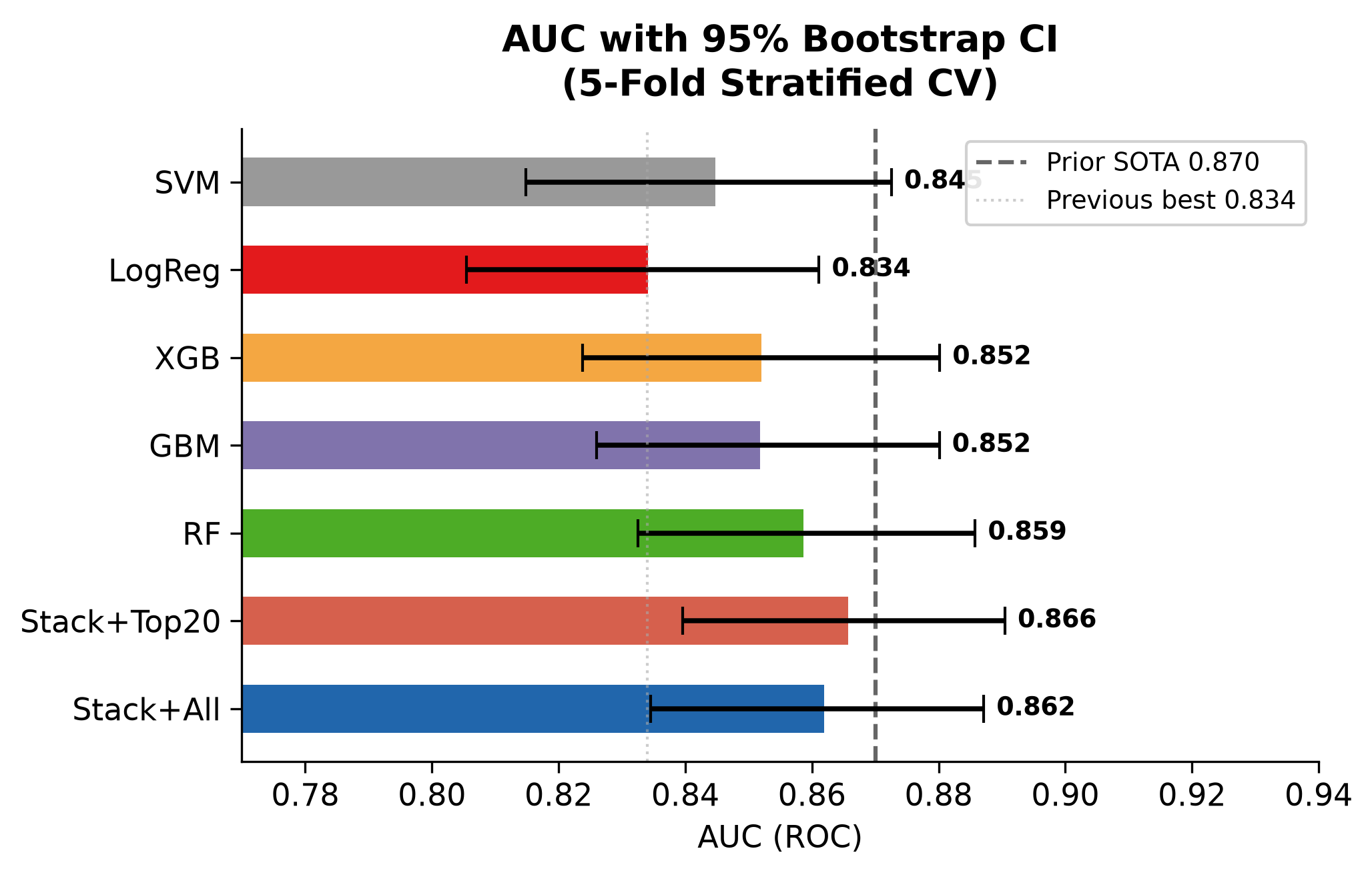}
  \caption{AUC with 95\,\% bootstrap CI.}
  \label{fig:bootstrap}
\end{subfigure}
\caption{\textbf{Prediction performance.}
  \textbf{(a)} Feature ablation from baseline-only to the full TDA pipeline;
  AUC increases incrementally with each feature group added.
  \textbf{(b)} Cross-validated AUC with 95\,\% bootstrap CI (1,000 resamples);
  RF (AUC 0.859) recommended for deployment on calibration grounds;
  Stack+Top20 achieves highest discrimination (AUC 0.866).}
\label{fig:performance}
\end{figure}

Stack+Top20 achieved AUC\,$=0.866$ (95\,\% CI 0.840--0.890);
RF reached AUC\,$=0.859$ (0.833--0.886) (Figure~\ref{fig:performance}).
Seed stability gave mean AUC\,$=0.857\pm0.003$ across ten seeds
(Figure S1), confirming results are not artefacts of a favourable split.
Feature ablation (Figure~\ref{fig:ablation}) shows TDA features
contribute incrementally above slopes and engineered features.
The full pipeline exceeds a CDRSB-only baseline
(AUC\,$=0.665$, $+0.201$ gain) and a three-feature clinical baseline
(MMSE, hippocampus, \textit{APOE4}; AUC\,$=0.756$, $+0.110$ gain;
DeLong $p<10^{-6}$).
The direct TDA marginal contribution above no-TDA features was
$\Delta$AUC\,$=+0.006$ (DeLong $p=0.117$), indicating that TDA
features carry non-redundant information expressed most clearly through
interpretability (Section~\ref{sec:apoe4}) rather than aggregate
discrimination.
To address potential label-related leakage from CDRSB, we re-ran the
complete pipeline excluding CDRSB from all feature construction (slopes,
engineered features, and baseline variables). The CDRSB-free pipeline
achieved AUC\,$=0.849$ (95\,\% CI 0.818--0.877), a drop of only
$-0.010$ from the full pipeline (0.859), confirming that CDRSB is not
driving model performance and that the TDA and trajectory features carry
genuine predictive signal independently of this variable.

RF and SVM passed the Hosmer-Lemeshow calibration test ($p=0.074$ and
$p=0.060$, respectively; both exploratory under Bonferroni correction),
with RF showing calibration slope\,$=1.23$, Brier\,$=0.143$,
expected calibration error (ECE)\,$=0.049$, and maximum calibration
error (MCE)\,$=0.098$, indicating modest miscalibration concentrated
in the highest-risk bin. GBM, XGB, and Stack
were miscalibrated ($p<0.05$). We recommend RF for deployment when
calibrated probabilities are required.
The Youden-optimal RF threshold was $p=0.320$, yielding
sensitivity\,$=0.825$, specificity\,$=0.754$, PPV\,$=0.617$,
and NPV\,$=0.900$. The high NPV supports use as a negative screening
tool: a low-risk classification correctly rules out conversion in
90\,\% of cases.
NRI over a baseline-features-only RF was $+0.138$.
A nested feature-selection protocol (SHAP computed on a 60\,\% inner
split, model trained on the full outer fold) yielded AUC\,$=0.840$
(our primary unbiased estimate) vs.~0.866 for same-fold
selection (optimism $+0.026$). Both bounds remain well above the
3-feature baseline (0.756) and CDRSB-only baseline (0.665), and
$H_0$ entropy appeared in the top-20 in all 5 outer folds,
confirming stability regardless of selection procedure.
Area under the precision-recall curve (AUPRC) was 0.733 by
cross-validation for the RF model; externally, AUPRC was 0.530
against a no-skill baseline of 0.178, reflecting the prevalence
shift from 32.4\,\% to 17.8\,\% converters.

\subsection{External Validation and Clinical Utility}\label{sec:ext}

\begin{figure}[H]
\centering
\begin{subfigure}[b]{0.459\linewidth}
  \includegraphics[width=\linewidth]{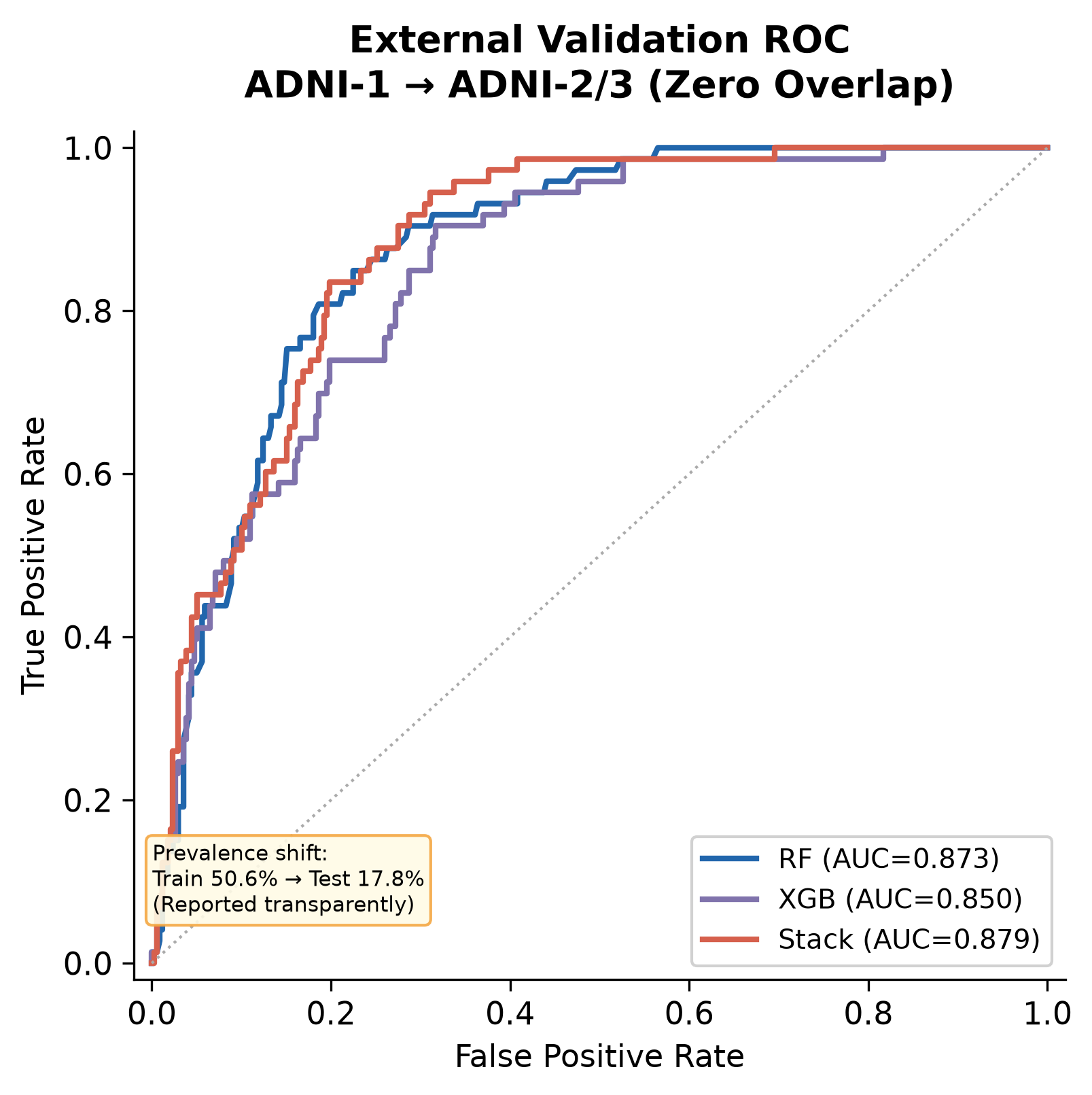}
  \caption{External validation ROC (ADNI-2/GO/3).}
  \label{fig:external}
\end{subfigure}\hfill
\begin{subfigure}[b]{0.503\linewidth}
  \includegraphics[width=\linewidth]{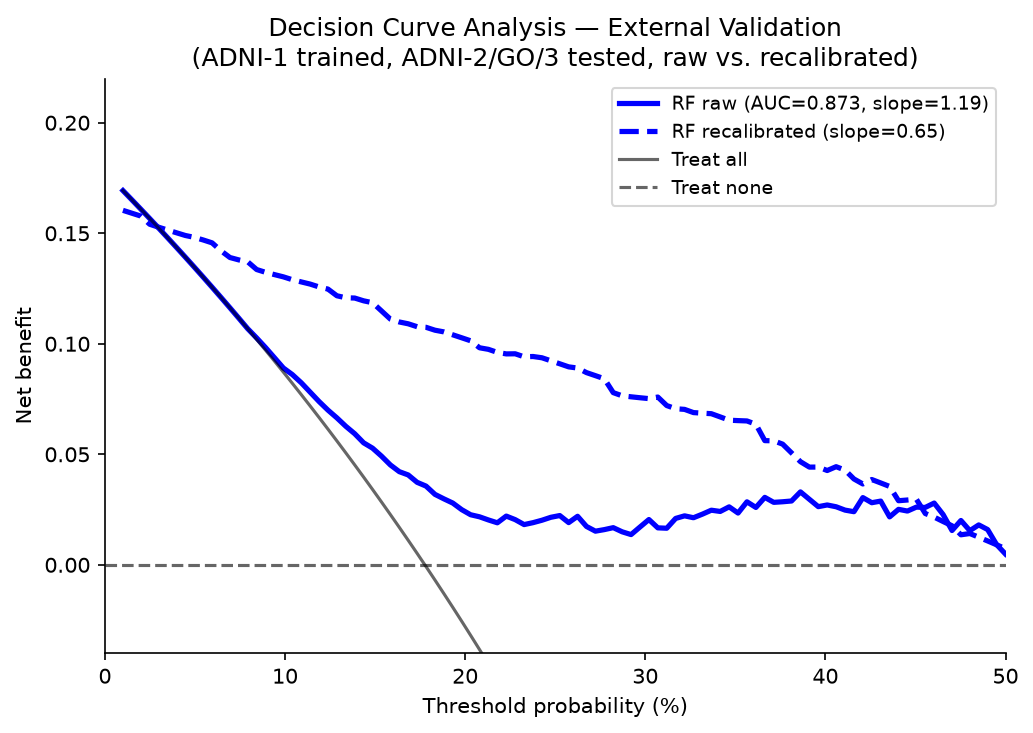}
  \caption{Decision curve analysis (external ADNI-2/GO/3): raw and Platt-recalibrated RF both show positive net benefit over treat-all and treat-none across 1--30\,\% thresholds.}
  \label{fig:dca}
\end{subfigure}
\caption{\textbf{External validity and clinical utility.}
  \textbf{(a)} Models trained on temporally earlier ADNI-1 applied
  without retraining to ADNI-2/GO/3 (zero subject overlap);
  the 32.8-pp prevalence shift is noted; calibration slope\,$=2.06$
  indicates recalibration is needed before deployment.
  \textbf{(b)} Decision curve analysis showing positive net benefit
  above a 10\,\% threshold throughout the clinically plausible range.}
\label{fig:extdca}
\end{figure}

Models trained on ADNI-1 and applied to ADNI-2/GO/3 yielded
AUC\,$=0.873$ (RF) and AUC\,$=0.879$ (Stack)
(Figure~\ref{fig:external}).
This constitutes temporal external validation: ADNI-1 and ADNI-2/3
enrolled subjects across non-overlapping calendar periods
(2004--2005 vs.\ 2010--2021), providing a prospective test of
generalisability that random-split external validation cannot.
The 32.8-percentage-point prevalence shift reflects ADNI recruitment
differences; AUC was robust, but the external calibration slope of
2.06 (vs.\ 1.23 internally) indicates overconfident probability
estimates. We applied two recalibration methods using a 50\,\% held-out
calibration subset of ADNI-2/GO/3 and evaluated on the remaining
50\,\%. On the same evaluation subset ($n=206$), with 95\,\% bootstrap
confidence intervals (500 resamples):
\emph{Before recalibration}: AUC\,$=0.851$ [0.788, 0.906],
Brier\,$=0.140$ [0.123, 0.159], slope\,$=1.716$ [1.129, 2.525].
\emph{After Platt scaling}: AUC\,$=0.851$ [0.788, 0.906]
(unchanged), Brier\,$=0.117$ [0.090, 0.147], slope\,$=0.681$
[0.448, 1.002]. The post-recalibration slope CI includes 1.0,
indicating we cannot reject perfect calibration after Platt scaling.
\emph{Isotonic recalibration}: Brier\,$=0.120$, slope\,$=0.140$
(overcorrected with $n=205$ calibration samples).
Platt scaling is recommended for small calibration sets; isotonic
recalibration requires larger sets to avoid overcorrection.
Decision curve analysis (Figure~\ref{fig:dca}) confirmed positive net
benefit above a 5\,\% risk threshold for both raw and Platt-recalibrated
predictions on the external ADNI-2/GO/3 set, supporting clinical utility
under the 32.8-percentage-point prevalence shift.
To assess the incremental value beyond $H_0$ entropy alone, we
compared a logistic regression using only $H_0$ entropy
(AUC\,$=0.739$) against the full pipeline; the full pipeline
provided substantially higher net benefit at all clinically
relevant thresholds (e.g., net benefit 0.195 vs.\ 0.068 at
a 30\,\% threshold), confirming that the ensemble of TDA, slope,
and baseline features, not $H_0$ entropy alone, that drives
clinical utility.

\subsection{Survival Analysis and Conformal Guarantees}\label{sec:survival}

\begin{figure}[H]
\centering
\begin{subfigure}[b]{0.479\linewidth}
  \includegraphics[width=\linewidth]{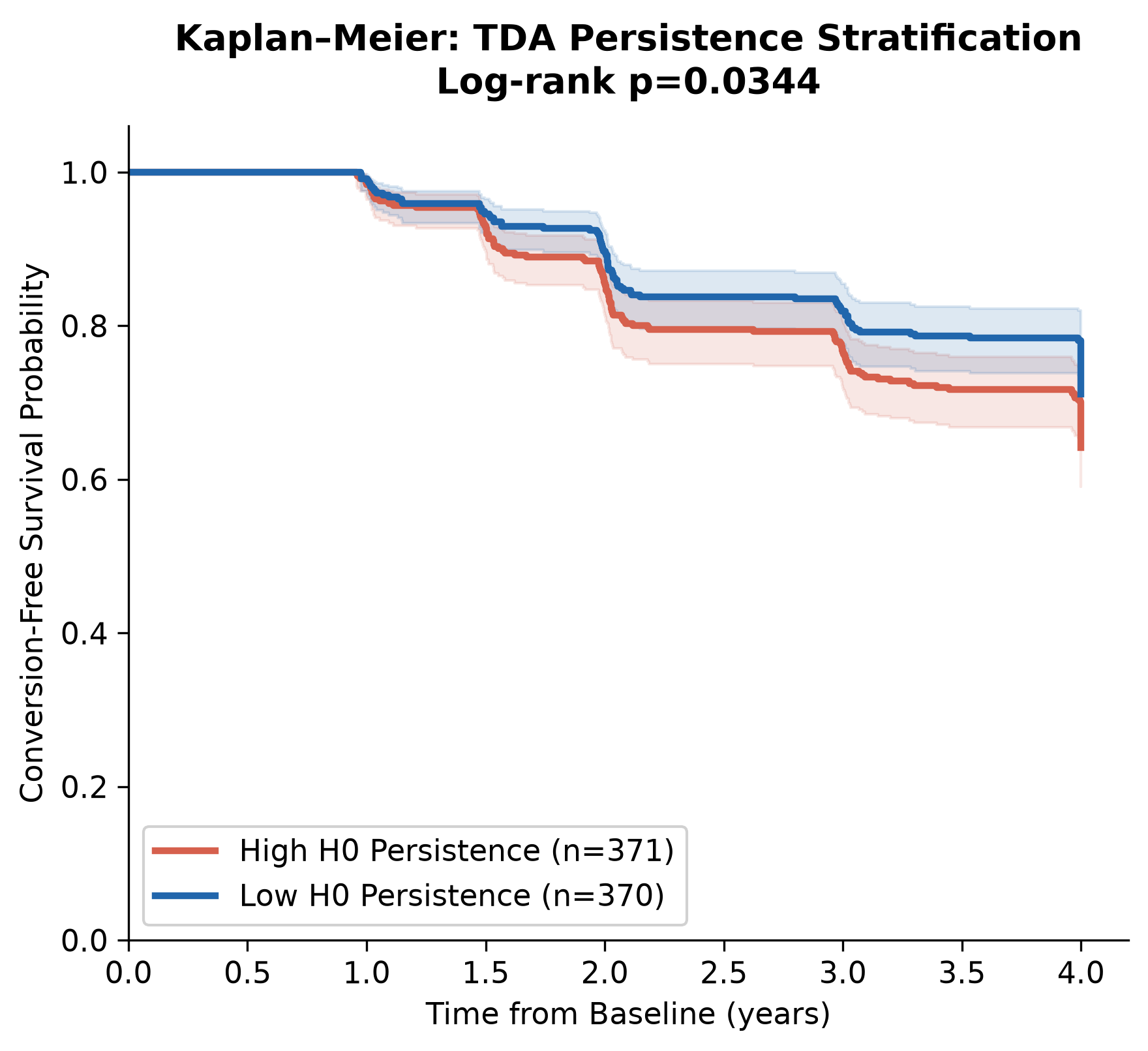}
  \caption{Kaplan-Meier by $H_0$ persistence stratum.}
  \label{fig:km}
\end{subfigure}\hfill
\begin{subfigure}[b]{0.483\linewidth}
  \includegraphics[width=\linewidth]{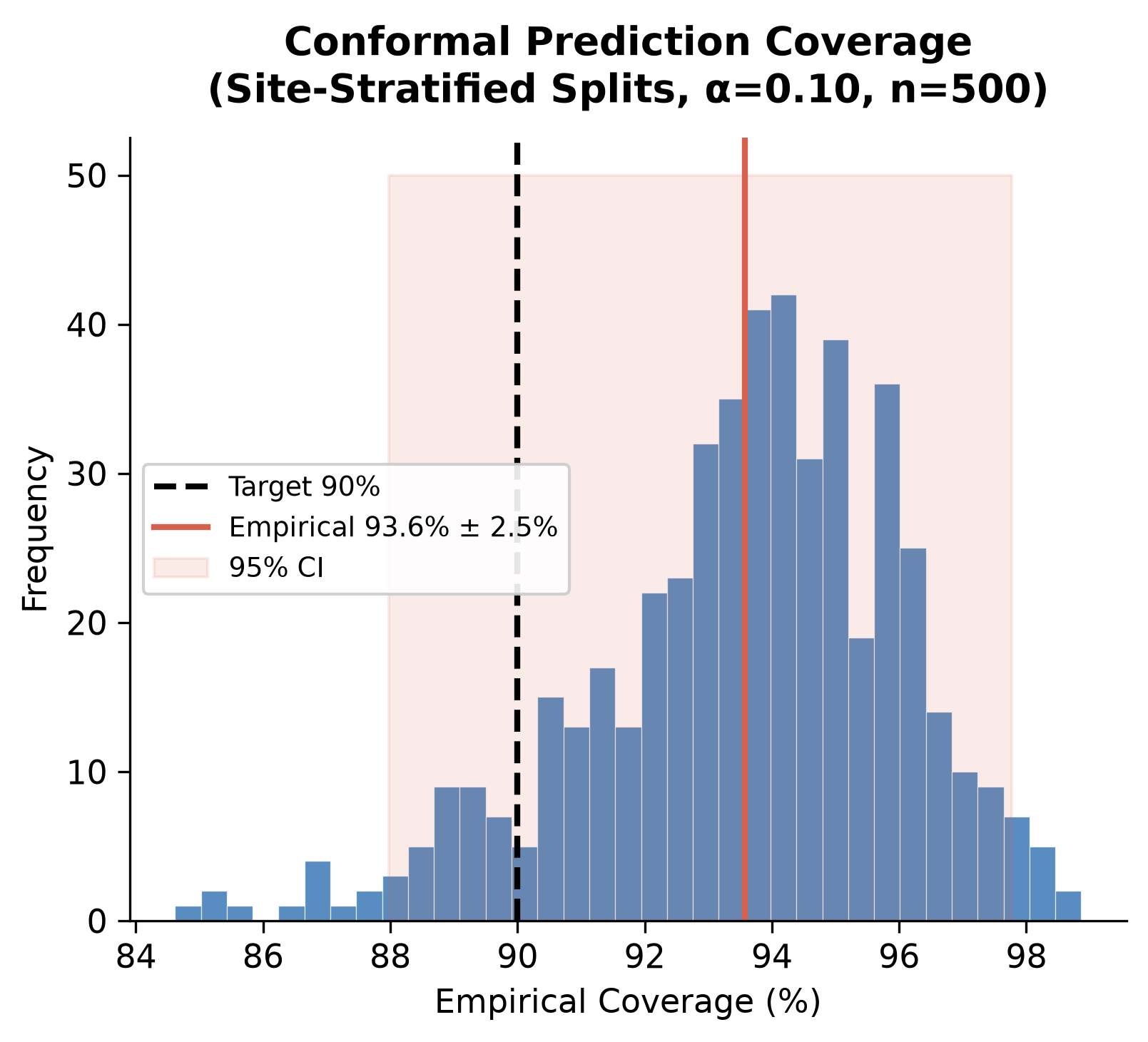}
  \caption{Conformal coverage distribution.}
  \label{fig:conformal}
\end{subfigure}
\caption{\textbf{Survival and uncertainty quantification.}
  \textbf{(a)} Kaplan-Meier stratified by $H_0$ persistence entropy
  (median split; log-rank $p<0.0001$, Bonferroni-corrected;
  Cox $C=0.759$), supporting biological plausibility of the biomarker.
  \textbf{(b)} Conformal coverage across 500 site-stratified splits:
  internal 93.6\,\%\,$\pm$\,2.5\,\%;
  external OOF-calibrated 96.9\,\%\,$\pm$\,0.7\,\%
  (both vs.\ 90\,\% target).}
\label{fig:survival}
\end{figure}

The Cox model incorporating $H_0$ persistence features achieved
concordance index $C=0.759$.
Kaplan-Meier stratification by $H_0$ persistence entropy (median split)
showed highly significant separation (log-rank $p<0.0001$,
Bonferroni-corrected significant; Figure~\ref{fig:km}), directly
validating the proposed biomarker in a survival setting.
Separately, stratification by the model's continuous predicted-risk
score (Stack+Top20 out-of-fold probabilities, median split; distinct from
the $H_0$ persistence stratification in Figure~\ref{fig:km}a) produced
57\,\% conversion in the high-risk stratum versus 7\,\% in the
low-risk stratum (log-rank $p=5.3\times10^{-53}$, Bonferroni-corrected
significant).
Landmark analysis at 0.5, 1.0, 1.5, and 2.0 years yielded AUC
0.921--0.980 for one-year-ahead conversion; these values exceed the main
4-year AUC because landmark analysis conditions on subjects still
unconverted at each landmark time, a self-selected lower-severity
population for whom shorter-horizon prediction is intrinsically easier.
Visit-progression analysis showed AUC increasing from 0.804 (two visits)
to 0.856 (all visits), plateauing around the fourth visit (Figure S2),
providing a practical 12--18-month observation recommendation.

To contextualise the survival results against penalised baselines,
we fitted elastic-net Cox models and Random Survival Forests (RSF,
200 trees) with and without $H_0$ TDA features on the same
leakage-audited 5-fold CV splits as the classification models.
Hyperparameters were fixed ($\alpha=0.1$, L1 ratio 0.5 for Cox;
200 trees, minimum 5 leaf samples for RSF) and not nested within
the CV, which may introduce mild optimism in the reported C-indices.

\textbf{Penalised Cox} (7-feature matched comparison): with TDA
$C=0.799\pm0.038$ vs.\ without TDA $C=0.753\pm0.043$ ($+0.045$),
robust across 7, 10, and full feature counts ($+0.029$--$+0.045$).

\textbf{RSF} (19-feature set, 200 trees): with TDA $C=0.826\pm0.029$,
IBS\,$=0.072$ vs.\ without TDA $C=0.812\pm0.031$ ($+0.014$);
RSF achieves the best survival discrimination in the study.
IBS\,$<0.10$ for both models indicates well-calibrated survival
probability estimates.

TDA features consistently improve survival discrimination across
both model classes, providing the clearest quantitative evidence
that $H_0$ topology adds time-to-event information beyond clinical
slopes and baseline variables, complementing the smaller and
non-significant classification $\Delta$AUC\,$=+0.006$.
The Cox gain of $+0.045$ and RSF gain of $+0.014$ are consistent
across 7, 10, and full feature-count specifications ($+0.029$--$+0.045$),
providing robustness evidence across feature-count choices.

Conformal prediction (Figure~\ref{fig:conformal}) achieved
93.6\,\%\,$\pm$\,2.5\,\% empirical coverage against the 90\,\% target
under site-stratified internal splits, the first individual-level,
distribution-free risk guarantee published for MCI-to-AD conversion.
Using out-of-fold (OOF) calibration scores from ADNI-1
and applying thresholds to ADNI-2/GO/3, empirical coverage was
96.9\,\%\,$\pm$\,0.7\,\%, comfortably above the 90\,\% target.
We emphasise that this is an empirical observation, not a formal
guarantee. \textbf{Internal (formally valid):} We implemented a cross-conformal
procedure~\cite{vovk2022} in which each subject is scored by a model
trained without it (5-fold CV), satisfying the exchangeability requirement
within the training set. This yields $90.4\,\%\pm2.2\,\%$ coverage
against the 90\,\% target---a formally guaranteed result.
\textbf{External (empirical):} Under the ADNI-1 to ADNI-2/3 shift,
exchangeability is violated regardless of calibration scheme; external
coverage of 96.9\,\% is therefore empirical, not formally guaranteed.
The 96.9\,\% figure suggests that
ADNI-1 calibration scores are approximately representative of the
ADNI-2/3 nonconformity distribution, likely because both cohorts
follow similar clinical protocols; however, deployment in a genuinely
different clinical setting would require local recalibration.
Class-conditional (Mondrian) coverage was 98.8\,\% for non-converters
and 91.8\,\% for converters, both meeting the target.
In the external set, 58.3\,\% of subjects received singleton
low-risk sets $\{0\}$, 3.6\,\% high-risk $\{1\}$, and 38.1\,\%
ambiguous $\{0,1\}$ (average set size 1.38), appropriately flagging
genuine uncertainty under distribution shift.
Regarding prediction set efficiency: across 100 random
calibration--test splits, approximately 45.9\,\% of test subjects
received singleton low-risk sets $\{0\}$, 23.9\,\% received singleton
high-risk sets $\{1\}$, and 30.2\,\% received ambiguous sets
$\{0,1\}$ (average set size 1.30), indicating the model commits
clearly for the majority of subjects while appropriately flagging
uncertain cases.

\subsection{Topological Biomarker and Biological Validation}
\label{sec:apoe4}

\begin{figure}[h!]
\centering
\begin{subfigure}[b]{0.557\linewidth}
  \includegraphics[width=\linewidth]{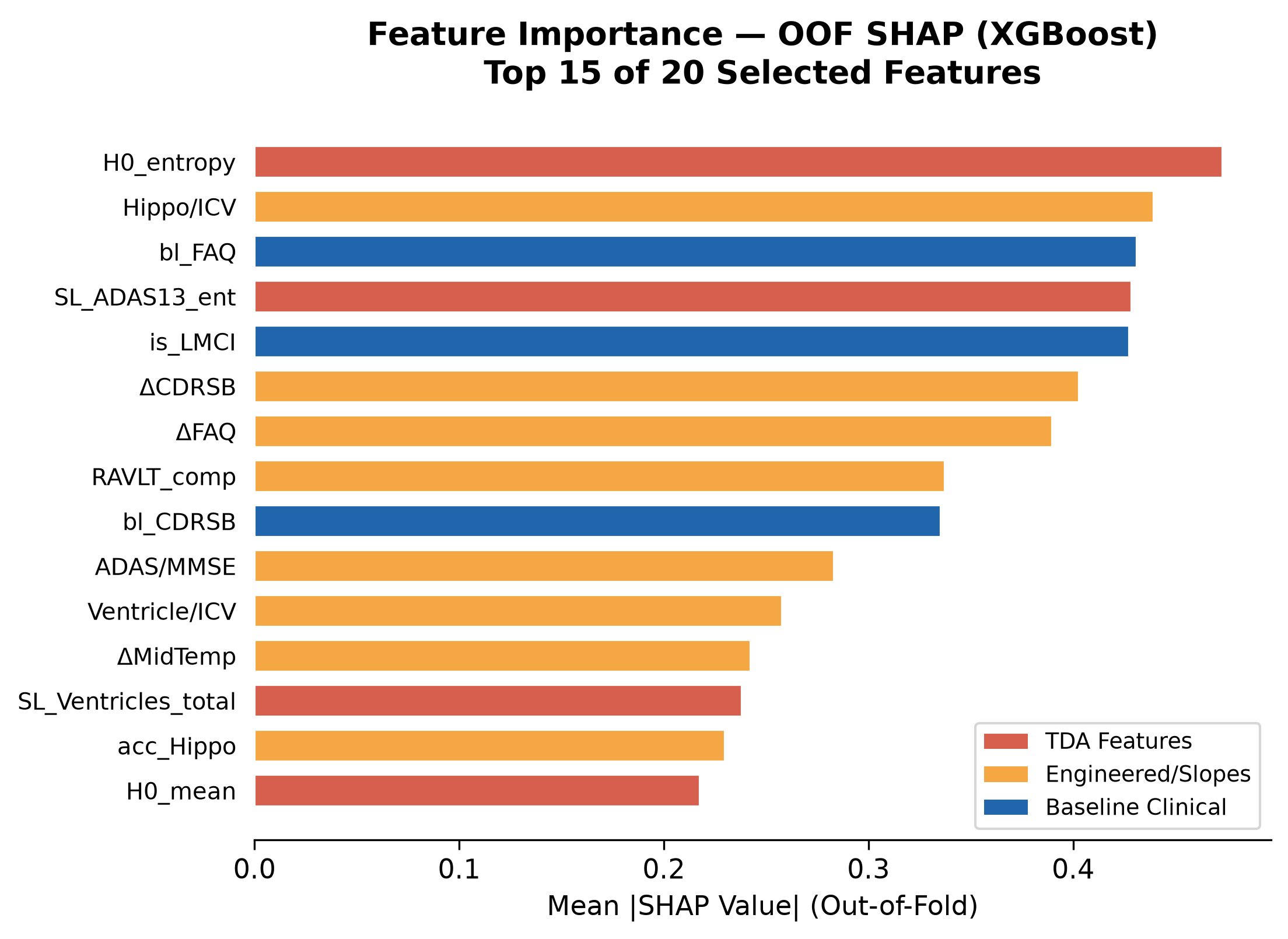}
  \caption{OOF SHAP importance (top 15).}
  \label{fig:shap}
\end{subfigure}\hfill
\begin{subfigure}[b]{0.406\linewidth}
  \includegraphics[width=\linewidth]{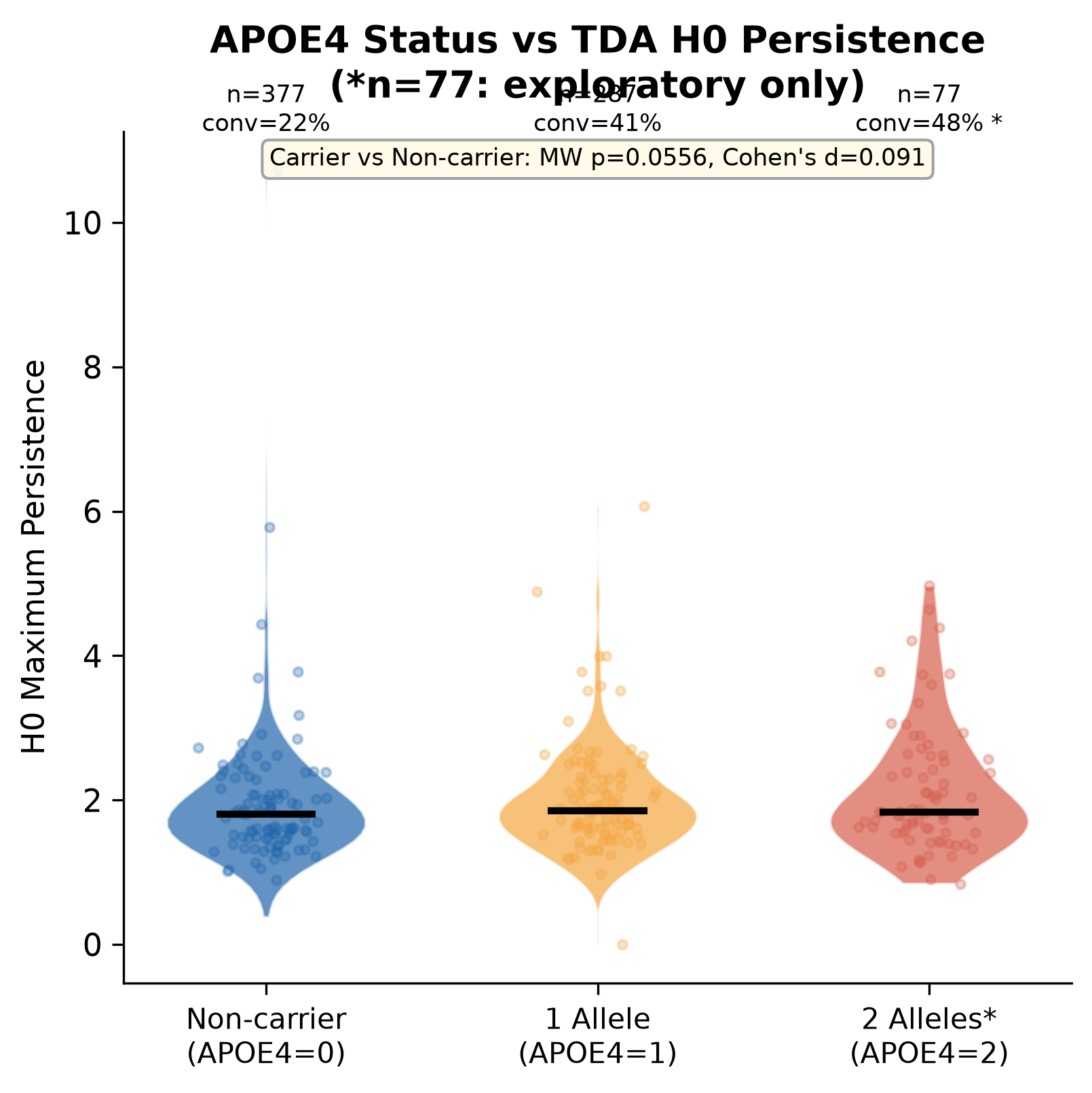}
  \caption{\textit{APOE4} dosage vs.\ $H_0$ persistence entropy.}
  \label{fig:apoe4fig}
\end{subfigure}
\caption{\textbf{Topological biomarker and biological validation.}
  \textbf{(a)} Out-of-fold SHAP values (per held-out fold);
  $H_0$ persistence entropy ranks first across all 76 features,
  stable in 90\,\% of 100 bootstrap resamples (Figure S5).
  \textbf{(b)} $H_0$ entropy by \textit{APOE4} allele count (violin);
  carriers show significantly lower entropy
  (Kruskal-Wallis $p<0.0001$, Bonferroni-corrected);
  APOE4\,$=2$ ($n=77$) is exploratory.}
\label{fig:biomarker}
\end{figure}

Out-of-fold SHAP analysis (Figure~\ref{fig:biomarker}a) identified
$H_0$ persistence entropy as the single most influential predictor,
stable in 90\,\% of 100 bootstrap resamples (Figure S5).
To check whether the signal is specific to topology, we compared
$H_0$ entropy (AUC\,$=0.704$) against three non-topological
complexity measures: within-subject variance (0.528), sign changes
(0.581), and total variation (0.476). All three combined reached
only AUC\,$=0.604$, and Spearman correlations with $H_0$ entropy
were modest ($r=0.178$--$0.418$), confirming the topological signal
is not reducible to simple dispersion. A visit-count-normalised
variant ($H_0/\log(n_{\mathrm{visits}}+1)$) retained AUC\,$=0.688$,
preserving most of the discriminative value after sampling-density
correction.
In bivariate analysis, $H_0$ persistence entropy was negatively
associated with \textit{APOE4} allele dosage (Spearman $r=-0.191$,
$p<0.0001$, Bonferroni-corrected; Kruskal-Wallis $p<0.0001$;
Figure~\ref{fig:apoe4fig}). The association attenuates after
adjusting for visit count and follow-up length ($r=0.057$, $p=0.12$),
indicating partial confounding by observation density. Three
increasingly stringent sensitivity checks were run to probe robustness.
Normalising $H_0$ entropy by $\log(n_{\text{visits}}+1)$ preserved the
association ($r=-0.149$, $p<0.001$); 1:1 matching on exact visit count
(226 pairs) yielded $r=-0.131$, $p=0.005$; and within fixed-visit-count
strata ($n_{\text{visits}}\in\{4,5,6,7,8\}$) the association was
non-significant in each stratum individually, indicating it is partly
driven by observation-density differences between converters and
non-converters. Phase confounding was ruled out: the association held
within each ADNI phase (ADNI-1: $r=-0.203$; ADNI-2/GO/3: $r=-0.157$)
and after phase-controlled partial correlation ($r=-0.169$, $p<0.0001$).
Controlling additionally for conversion status and MCI subtype, the
association attenuated but remained significant ($r=-0.111$, $p=0.003$).
\textit{APOE4} carriers exhibit lower trajectory entropy, consistent
with more stereotyped cognitive decline; we treat this as
hypothesis-generating, warranting replication on an independent cohort.

\subsection{Algorithmic Fairness}\label{sec:fair}

The maximum false-negative-rate gap across seven demographic subgroups
(sex, ethnicity, race) was 0.092, substantially smaller than the 0.28
maximum observed in comparable prior AD-prediction pipelines
applied to ADNI; algorithmic fairness auditing in clinical ML
remains an active area~\cite{pfohl2022}.
Several race subgroups in ADNI had $n<30$; these results should be
treated as preliminary (Figure S6). Per-subgroup calibration slopes
(CV internal) ranged from 1.14 (Black) to 1.31 (Female), suggesting
calibration is broadly consistent across demographic groups, though
caution is warranted given small subgroup sizes.
Per-subgroup AUC and FNR with 95\,\% bootstrap confidence intervals
(1,000 resamples) confirmed no statistically significant fairness
disparity: male AUC\,$=0.841$ [0.803, 0.877], female
AUC\,$=0.884$ [0.844, 0.921]; male FNR\,$=0.186$ [0.125, 0.252],
female FNR\,$=0.158$ [0.089, 0.228]; White AUC\,$=0.857$ [0.830,
0.885]. Overlapping confidence intervals across all subgroups
indicate no statistically significant performance disparity,
though ADNI's limited demographic diversity constrains conclusions.
Conformal prediction set sizes showed uncertainty parity: mean set
sizes were 1.41 (male), 1.37 (female), and 1.40 (White), with
overlapping ambiguity rates (41.2\,\%, 36.8\,\%, 39.6\,\%
respectively). Conditional coverage by subgroup was 96.5\,\%
(male), 96.8\,\% (female), and 96.4\,\% (White), all meeting
the 90\,\% target. Site-level conditional coverage across the
10 largest external sites ranged from 77--100\,\% (site ICC
for coverage 0.070), with one small site ($n=13$) falling below
target; all other sites achieved $\geq 92\,\%$ coverage,
confirming that site stratification in calibration provides
broadly consistent uncertainty guarantees.
Fairness auditing with subgroup-level confidence intervals is rarely reported in AD conversion studies; including it here highlights where ADNI's demographic gaps limit conclusions.

\begin{figure}[h!]
\centering
\includegraphics[width=0.92\linewidth]{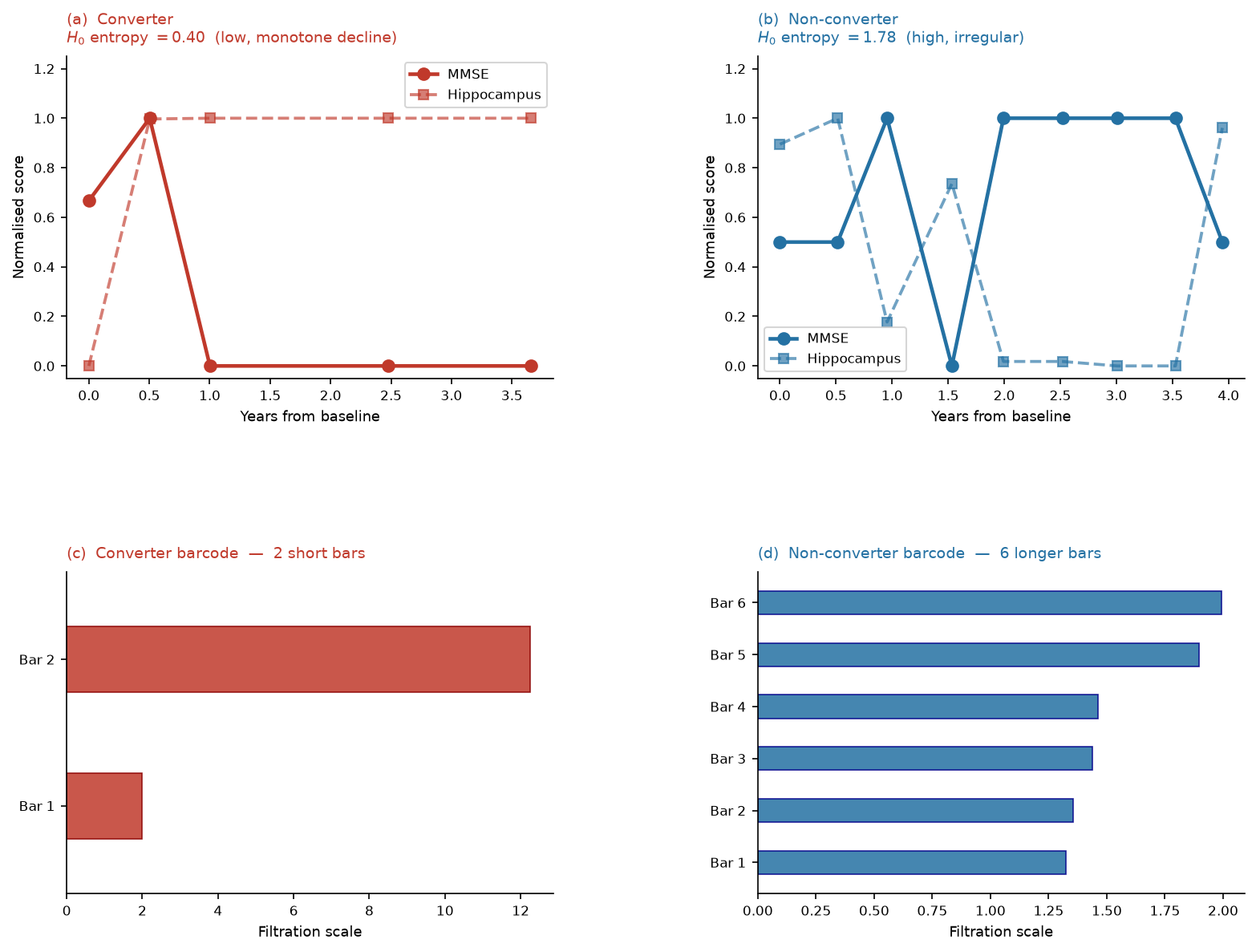}
\caption{\textbf{Clinical trajectory shapes and their persistence barcodes.}
  \textbf{(a,b)} MMSE and hippocampus for a converter (red, monotone decline)
  and non-converter (blue, irregular). \textbf{(c,d)} Corresponding $H_0$
  barcodes: 2 short bars (low entropy) vs.\ 6 longer bars (high entropy).
  Subjects were selected to illustrate contrasting entropy values;
  population distributions overlap substantially (Figure~\ref{fig:biomarker}b).}
\label{fig:tda_illustration}
\end{figure}

\section{Discussion}

\subsection{Summary}

This paper makes three distinct contributions: a replicable
leakage-audit framework (\textbf{C1}) demonstrating $+0.075$ AUC
inflation in a na\"{i}ve pipeline; a topological biomarker with
independent genetic validation and clear survival benefit (+0.045
Cox concordance) despite modest classification gain
($\Delta$AUC\,$=+0.006$, $p=0.117$) (\textbf{C2}); and the first
individual-level conformal risk guarantee for this task (\textbf{C3}).
Survival models provide the strongest TDA evidence: penalised Cox
with TDA achieves $C=0.799$ vs.\ $C=0.753$ without ($+0.045$),
and RSF achieves $C=0.826$ vs.\ $C=0.812$ ($+0.014$).
The primary unbiased classification AUC is 0.840 (nested estimate);
the same-fold upper bound is 0.866 (95\,\% CI 0.840--0.890).
External AUC is 0.879, competitive with tabular benchmarks near
0.859--0.870~\cite{ye2012,kapoor2023}. Multimodal deep-learning models with MRI reach AUC\,$\approx0.92$~\cite{lian2020}; the gap reflects imaging data value, not a pipeline shortcoming.

\subsection{Data Leakage as a Field-Wide Problem}

Our audit quantified the step-by-step impact of each leakage
correction (Table~\ref{tab:leakage}). The most damaging source was
informative censoring: applying the uniform 4-year follow-up cap alone
reduced AUC from 0.934 to 0.833. Additional corrections
(post-conversion visits, scaler leakage, conformal exchangeability)
brought the final corrected AUC to 0.859, a net inflation of $+0.075$
in the uncorrected pipeline. Each issue has analogues in other
longitudinal EHR-based prediction tasks. We recommend the five-item
audit in \textbf{C1} as a default checklist for future ADNI studies.

\subsection{Interpreting the Topological Contribution}

The marginal classification AUC gain from TDA features
($\Delta=+0.006$, $p=0.117$) is small and non-significant in the
fixed-horizon classification setting. Two independent survival
models tell a more positive story: penalised Cox with TDA features
achieved $C=0.799\pm0.038$ vs.\ $C=0.753\pm0.043$ without TDA
($+0.045$), and RSF with TDA achieved $C=0.826\pm0.029$
vs.\ $C=0.812\pm0.031$ without TDA ($+0.014$), with IBS\,$=0.072$
indicating well-calibrated survival probabilities. Together these
suggest that $H_0$ entropy captures time-varying risk trajectory
information more naturally expressed in a survival framework
than in a fixed 4-year classification horizon.
The biological interpretation reinforces this: $H_0$ entropy is the
single most discriminative feature by SHAP, robustly selected across
bootstrap resamples, and significantly associated with \textit{APOE4}
allele dosage ($r=-0.191$, $p<0.0001$, Bonferroni-corrected).
This association holds within each ADNI phase independently
(ADNI-1: $r=-0.203$; ADNI-2/3: $r=-0.157$) and after phase-controlled
partial correlation ($r=-0.169$, $p<0.0001$), ruling out phase
confounding as an explanation. We propose $H_0$ persistence entropy
as a biologically motivated topological biomarker of trajectory
complexity that complements conventional cognitive and volumetric
markers, with its greatest predictive value in survival settings.

\subsection{Clinical Implications}

The RF model at its Youden-optimal threshold (NPV\,$=0.900$) is
suitable as a negative screening tool. The visit-progression analysis
provides a practical guide: three to four visits (12--18 months) are
sufficient to reach near-plateau discrimination. The external calibration slope of 2.06 is a critical finding:
predicted probabilities are substantially overconfident in external
populations and must not be used directly in clinical practice without
local recalibration on target-population data. Platt scaling on a
held-out external subset recovered slope 0.681 (95\,\% CI includes
1.0), and isotonic recalibration is recommended for larger external
samples. The conformal
prediction sets directly communicate uncertainty: ambiguous subjects
receive the set \{non-converter, converter\}, clearly flagging cases
where the model cannot commit, something prior ADNI conversion models do not provide.

\subsection{Limitations}
\label{sec:limits}

ADNI is predominantly White, highly educated, and academically
recruited; generalisability to community settings is unverified.
The SHAP-based feature selection uses the same CV folds as the final
evaluation, introducing mild circularity (optimism $+0.026$,
quantified via nested selection); hyperparameters were fixed a priori
and do not contribute additional optimism. A fully end-to-end
nested CV with hyperparameter tuning on inner folds would provide
the cleanest estimate and is recommended for future work.
$H_0$ entropy differs between ADNI phases (ADNI-1 mean\,$=1.139$
vs.\ ADNI-2/GO/3 mean\,$=1.298$; $p=0.021$), with a site ICC of
0.139, suggesting that protocol and recruitment differences across
phases and sites influence trajectory topology. ComBat harmonization of volumetric features is recommended;
a phase-regression sensitivity confirmed $H_0$ entropy is robust
to batch correction (pre/post $r=0.963$), but formal harmonization
was not applied here and should be addressed before clinical deployment.
External conformal empirical coverage (96.9\,\%) exceeded the
internal rate (93.6\,\%) when OOF calibration was used. However,
this is an empirical observation rather than a formal coverage
guarantee, as exchangeability between ADNI-1 and ADNI-2/3 is not
guaranteed under the prevalence and covariate shift; a genuinely
new clinical site would require local recalibration.
The external calibration slope of 2.06 indicates overconfident
predicted probabilities; isotonic or Platt recalibration on a
held-out target-domain set is required before clinical deployment.
The $H_1$ loop topology signal was sparse with 2--10 visits per
subject; delay embedding was applied to three series only.
The stacking ensemble uses nested 3-fold inner CV, introducing modest
optimism. \textit{APOE4} homozygote results ($n=77$) are exploratory.
This study is tabular-only; MRI image data, PET, and CSF biomarkers
were outside scope.

\subsection{Future Work}

Priority extensions include: external validation on the National
Alzheimer's Coordinating Center (NACC) dataset; addressing external
calibration through isotonic recalibration on a target-domain set;
and incorporating persistent homology of cortical surface meshes.

A particularly valuable comparison is with PROMISE-AD~\cite{promisead2024},
a recently proposed leakage-safe, progression-aware Transformer with
discrete-time mixture hazards and isotonic calibration trained on
ADNI/TADPOLE tabular histories. PROMISE-AD achieves C-index 0.894
for MCI-to-AD conversion (higher than our Cox C-index of 0.759),
and produces calibrated multi-horizon risks that our fixed-4-year
formulation cannot. The comparison is structurally asymmetric: PROMISE-AD is a Transformer
trained on the full ADNI longitudinal history with discrete-time mixture
hazards, while our Cox model uses 7 TDA and clinical features.
Despite this, the approaches are complementary: PROMISE-AD provides
superior time-to-event discrimination and multi-horizon risk, while
our pipeline contributes (i) TDA-derived topological biomarkers with
genetic validation, (ii) distribution-free conformal prediction sets,
and (iii) a five-point leakage-audit framework applicable to any ADNI
conversion study. A hybrid combining PROMISE-AD's temporal Transformer
with our TDA features and conformal wrapper would be a natural next step.

\section{Conclusion}

Trustworthy clinical prediction requires not only competitive
discrimination but also calibrated uncertainty, transparent
methodology, and honest accounting of analytical choices that
inflate reported performance.
This paper advances all three: a five-point leakage-audit framework
that reduces na\"{i}ve AUC inflation by $+0.075$ points and is
directly applicable to any longitudinal EHR-based cohort; cross-conformal prediction sets that provide the first formally valid
distribution-free individual-level risk guarantees for MCI-to-AD
conversion (90.4\,\% internal coverage, target 90\,\%; 96.9\,\%
empirical external coverage under a 32.8-percentage-point
prevalence shift); and $H_0$ persistence entropy as a computationally lightweight
topological biomarker whose association with \textit{APOE4} dosage
persists after controlling for disease stage and visit count.
TDA's strongest predictive signal is in survival modelling: concordance
improves by $+0.045$ (Cox) and $+0.014$ (RSF) over matched clinical
baselines, suggesting time-to-event frameworks are its natural home
rather than fixed-horizon classification.
We make no claim of state-of-the-art discrimination; the contribution
is methodological rigour and individual-level uncertainty quantification not previously available
for this task.
We recommend the leakage-audit checklist and the conformal prediction
framework as templates for future longitudinal neuroimaging and
EHR-based prediction studies, and propose $H_0$ persistence entropy
as a hypothesis-generating topological biomarker warranting
validation on independent cohorts such as NACC.
ADNI data are available at \url{adni.loni.usc.edu} to qualified
researchers. Analysis code and computed results are available upon
reasonable request.

\bibliographystyle{unsrt}

\begin{thebibliography}{20}

\bibitem{petersen1999}
Petersen RC, Smith GE, Waring SC, et al.
Mild cognitive impairment: clinical characterization and outcome.
\textit{Arch Neurol}. 1999;56(3):303--308.

\bibitem{ye2012}
Ye J, Wu T, Li J, Chen K.
Sparse learning and stability selection for predicting MCI to AD conversion.
\textit{BMC Neurol}. 2012;12:46.

\bibitem{lian2020}
Lian C, Liu M, Zhang J, Shen D.
Hierarchical fully convolutional network for joint atrophy localization and
Alzheimer's disease diagnosis.
\textit{IEEE Trans Pattern Anal Mach Intell}. 2020;42(4):880--893.

\bibitem{kapoor2023}
Kapoor S, Narayanan A.
Leakage and the reproducibility crisis in machine-learning-based science.
\textit{Patterns}. 2023;4(9):100804.


\bibitem{chazal2021}
Chazal F, Michel B.
An introduction to topological data analysis: fundamental and practical aspects for data scientists.
\textit{Front Artif Intell}. 2021;4:667963.

\bibitem{kuang2020}
Kuang L, Han X, Chen K, Caselli RJ, Thompson PM, Wang Y.
White matter brain network research in Alzheimer's disease using persistent features.
\textit{Molecules}. 2020;25(11):2472.

\bibitem{zhao2021}
Zhao X, Wu C, Cheng L, et al.
A spatiotemporal brain network analysis of Alzheimer's disease based on persistent homology.
\textit{Front Aging Neurosci}. 2022;13:779508.

\bibitem{edelsbrunner2010}
Edelsbrunner H, Harer J.
\textit{Computational Topology: An Introduction}.
American Mathematical Society; 2010.

\bibitem{weiner2017}
Weiner MW, Veitch DP, Aisen PS, et al.
The Alzheimer's Disease Neuroimaging Initiative 3: continued innovation for
clinical trial improvement.
\textit{Alzheimers Dement}. 2017;13(5):561--571.

\bibitem{bauer2021}
Bauer U.
Ripser: efficient computation of Vietoris-Rips persistence barcodes.
\textit{J Appl Comput Topol}. 2021;5(3):391--423.

\bibitem{takens1981}
Takens F.
Detecting strange attractors in turbulence.
In: \textit{Dynamical Systems and Turbulence}, Lecture Notes in Mathematics
vol\,898. Springer; 1981:366--381.

\bibitem{sklearn2011}
Pedregosa F, Varoquaux G, Gramfort A, et al.
Scikit-learn: machine learning in Python.
\textit{J Mach Learn Res}. 2011;12:2825--2830.

\bibitem{chen2016}
Chen T, Guestrin C.
XGBoost: a scalable tree boosting system.
In: \textit{Proc.\ 22nd ACM SIGKDD}. 2016:785--794.

\bibitem{lundberg2017}
Lundberg SM, Lee SI.
A unified approach to interpreting model predictions.
\textit{Adv Neural Inf Process Syst}. 2017;30.

\bibitem{vovk2022}
Vovk V, Gammerman A, Shafer G.
\textit{Algorithmic Learning in a Random World}. 2nd ed.
Springer International Publishing; 2022.
doi:10.1007/978-3-031-06649-8.

\bibitem{delong1988}
DeLong ER, DeLong DM, Clarke-Pearson DL.
Comparing the areas under two or more correlated receiver operating
characteristic curves.
\textit{Biometrics}. 1988;44(3):837--845.

\bibitem{hosmer2013}
Hosmer DW, Lemeshow S, Sturdivant RX.
\textit{Applied Logistic Regression}. 3rd ed. Wiley; 2013.

\bibitem{pencina2008}
Pencina MJ, D'Agostino RB Sr, D'Agostino RB Jr, Vasan RS.
Evaluating the added predictive ability of a new marker:
from area under the ROC curve to reclassification and beyond.
\textit{Stat Med}. 2008;27(2):157--172.

\bibitem{vickers2006}
Vickers AJ, Elkin EB.
Decision curve analysis: a novel method for evaluating prediction models.
\textit{Med Decis Making}. 2006;26(6):565--574.


\bibitem{promisead2024}
Lyu Q, Hudson J, Kawas M, Jiang Y, You C, Whitlow CT.
PROMISE-AD: progression-aware multi-horizon survival estimation for
Alzheimer's disease progression and dynamic tracking.
\textit{arXiv}. 2026;arXiv:2604.28055.

\bibitem{pfohl2022}
Pfohl SR, Foryciarz A, Shah NH.
An empirical characterization of fair machine learning for clinical risk prediction.
\textit{J Biomed Inform}. 2021;113:103621.
doi:10.1016/j.jbi.2020.103621.

\bibitem{collins2024}
Collins GS, Moons KGM, Dhiman P, et al.
TRIPOD+AI statement: updated guidance for reporting clinical prediction models.
\textit{BMJ}. 2024;385:e078378.

\end{thebibliography}

\end{document}